\documentclass[twocolumn,conference]{IEEEtran}
\usepackage[T1]{fontenc}
\usepackage[latin9]{inputenc}
\usepackage{color}
\usepackage{verbatim}
\usepackage{mathrsfs}
\usepackage{amsmath}
\usepackage{amssymb}
\usepackage{graphicx}
\usepackage[unicode=true,
 bookmarks=true,bookmarksnumbered=true,bookmarksopen=true,bookmarksopenlevel=1,
 breaklinks=false,pdfborder={0 0 0},pdfborderstyle={},backref=false,colorlinks=false]
 {hyperref}
\hypersetup{pdftitle={Your Title},
 pdfauthor={Your Name},
 pdfpagelayout=OneColumn, pdfnewwindow=true, pdfstartview=XYZ, plainpages=false}

\makeatletter

\providecommand{\tabularnewline}{\\}

\usepackage[caption=false,font=footnotesize]{subfig}

\@ifundefined{showcaptionsetup}{}{%
 \PassOptionsToPackage{caption=false}{subfig}}
\usepackage{subfig}
\makeatother

\begin{document}
\title{Goal-driven Long-Term Trajectory Prediction}
\author{Hung Tran, Vuong Le, Truyen Tran\\
Applied AI Institute, Deakin University, Geelong, Australia \\
 \texttt{\small{}\{tduy,vuong.le,truyen.tran\}@deakin.edu.au}}
\maketitle
\begin{abstract}
The prediction of humans' short-term trajectories has advanced significantly
with the use of powerful sequential modeling and rich environment
feature extraction. However, long-term prediction is still a major
challenge for the current methods as the errors could accumulate along
the way. Indeed, consistent and stable prediction far to the end of
a trajectory inherently requires deeper analysis into the overall
structure of that trajectory, which is related to the pedestrian's
intention on the destination of the journey. In this work, we propose
to model a hypothetical process that determines pedestrians' goals
and the impact of such process on long-term future trajectories. We
design Goal-driven Trajectory Prediction model - a dual-channel neural
network that realizes such intuition. The two channels of the network
take their dedicated roles and collaborate to generate future trajectories.
Different than conventional goal-conditioned, planning-based methods,
the model architecture is designed to generalize the patterns and
work across different scenes with arbitrary geometrical and semantic
structures. The model is shown to outperform the state-of-the-art
in various settings, especially in large prediction horizons. This
result is another evidence for the effectiveness of adaptive structured
representation of visual and geometrical features in human behavior
analysis.

\end{abstract}

\global\long\def\GE{\text{Goal Encoder}}%
\global\long\def\GD{\text{Goal Decoder}}%
\global\long\def\TE{\text{Trajectory Encoder}}%
\global\long\def\TD{\text{Trajectory Decoder}}%

\global\long\def\ModelName{\text{Goal-driven Trajectory Prediction}}%
\global\long\def\Model{\text{GTP}}%

\section{Introduction }

\begin{figure}
\begin{centering}
\includegraphics[width=0.49\textwidth,height=8cm,keepaspectratio]{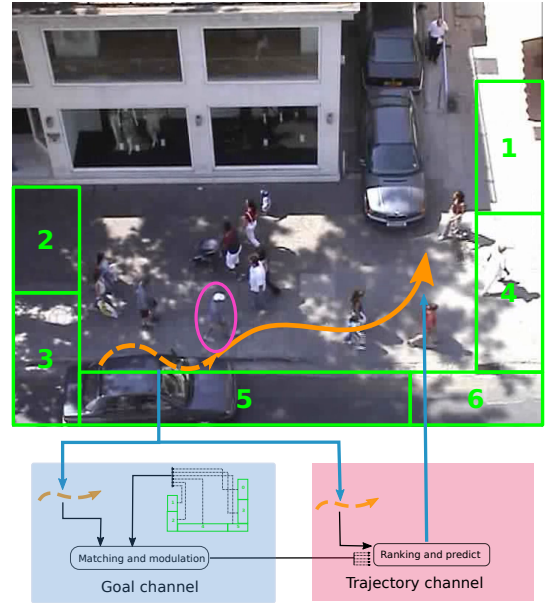}
\par\end{centering}
\caption{$\protect\ModelName$ model decomposes the movement of a pedestrian
into two concurrent sub-processes: Goal process governing the long-term
intention toward a destination, and Trajectory process controlling
detail movements. This dual process is implemented as a neural network
of two-channels that collaborate with each other to generate future
trajectory. Destinations are automatically identified as reachable
regions at the boundary of the scene. \label{fig:intro-demonstration}}
\end{figure}

The behavior of humans represented in their walking trajectories is
a complex process that provides a rich ground for mathematical and
machine modeling. There are two fundamental types of factors that
influence the behavior: Firstly, a pedestrian keeps an intention on
a destination they want to reach; and this goal governs the long-term
tendency of their trip. Secondly, along the way to the destination,
the pedestrian needs to make short-term adjustments according to immediate
situations such as the physical terrain and other moving agents. Understanding
and characterizing this dual process of intention and adjustment promise
effective coarse-to-fine trajectory modeling and hence improve prediction
performance.

Since the long-term intention is vague and difficult to model, available
studies on pedestrian trajectory prediction biased into learning the
short-term adjustment. This is usually done by exploiting the temporal
consistency of the trajectory under the assumption that the movement
pattern observed in the past extends to the future. Top-performing
methods utilized deep learning based sequential models such as variations
of Recurrent Neural Networks (RNNs) \cite{GRU,LSTM}. Most recent
developments in this problem focused on enriching the input of the
sequential models by features of the surrounding environment \cite{desire,peekingintofuture,sophie,spatially-aware-attention}
and social interaction with other agents \cite{sociallstm,socialgan,spatio-temporal-graph,social-attention,sr-lstm}.
Although these enhancements improved short-term prediction, the impact
fell short in long-term prediction because the pedestrian's goal is
unaccounted for \cite{trajectory-survey}.

In this work, we endeavor to explicitly model the dependency of pedestrians'
trajectories on their intention toward possible destinations. We hypothesize
that the navigation process of a pedestrian could be expressed by
two sup-processes: goal estimation and trajectory prediction. These
coupled sub-processes are modeled by a dual-channel neural network
named $\ModelName$ (abbreviated $\Model$, see Fig. \ref{fig:intro-demonstration}).
The model consists of two interactive sub-networks: \emph{the Goal
channel} estimates the intention of the subject toward the auto-selected
destinations and provides guidance for \emph{the Trajectory channel}
to predict details future movements. The interaction between the two
channels is done by a flexible attention mechanism on the provided
guidance of the Goal channel so that we could maintain the balance
between strong far-term planning and short-term trajectory adjustment.
In fact, the whole architecture design resembles the way the human
brain uses two biological neural sub-networks to control our attention:
top-down cognitive-related network and bottom-up stimulus reaction
network \cite{dual-process-attention}. Among the two, the former
shares conceptual similarities with our Goal channel, while the latter
is related to our Trajectory channel.

The destinations used in GTP are detected on-the-spot adaptively to
the semantic segmentation of the scene. The two channels of GTP are
trained to rank these flexible destinations through attention and
compatibility matching. This zero-shot mechanism supports transfer
learning to unseen scenes, which resolves the weakness of traditional
goal-based methods. 

In our cross-scene prediction experiments on ETH and UCY datasets,
we demonstrate the effectiveness of the Goal channel in the overall
planning, as the far-term prediction of our model improves significantly
compared to the current state-of-the-art. In the meantime, we also
showed the role of the Trajectory channel in considering both guidance
from the Goal channel and immediate circumstances for precise in-time
adjustment. 

Our method of utilizing goal information in future trajectory forecasting
is another step toward describing the natural structure of human behavior.
Also, the representation power of the ranking-based system enables
our model to generalize across unknown scenes and raise a new bar
in far-term trajectory prediction - the major challenge of the field.

\section{Related Work}

\textbf{Pedestrian trajectory prediction }recently achieved much improvement
with the deep learning approaches \cite{trajectory-survey}. By treating
human trajectory as time-series data, these approaches use variations
of Recurrent Neural Networks \cite{GRU,LSTM} to learn the temporal
continuity of the subjects' locations and movements. Beyond temporal
consistency, recent efforts concentrated on adding human-human interaction
by various methods for social pooling \cite{sociallstm,group-lstm,social_pooling2,socialgan,social_pooling1,social_pooling3}
and social-based refinement \cite{social_way,socialgan,desire,sophie,social-attention,sr-lstm}.
Aside from the dynamic social interaction, many efforts are spent
on examining the static environment surrounding the subjects that
may affect their trajectory. These environmental factors are extracted
either directly from the image \cite{desire,sophie,ss_lstm,scene_img1}
or from the semantic map of the environment \cite{peekingintofuture,social_pooling1,spatially-aware-attention}.
Although passive and active entities contribute significantly to the
immediate future behavior, their effects fade out in long term. By
contrast, we investigate how changing environmental context interacts
with the lasting end goal of the trajectory, which allows reliable
forecasting far into the future.

\textbf{Intention oriented trajectory prediction }has been approached
for robots and autonomous vehicles in the form of planning-based navigation
engines on top of a Markov decision process (MDP) solver \cite{intent-aware,activity-forcasting,goal-directed,planning_mdp1,joint-long-term-prediction,planning-based-prediction}.
The plans of vehicles are laid out on a grid-based map formed by discretizing
the scene. This simplification limits the generalization capacity
to scenes with different scales or configurations. Hence, several
methods instead used recurrent neural networks to work with continuous
representations of scenes and goals \cite{precog,paper-drogon,social_bigat,multipleprediction}.
In these approaches, the agent chooses one among a given set of goals
and plans the trajectory to accomplish it. The rigid plans used in
these methods are not readily applicable to pedestrian trajectories
because unlike vehicles that follow lane lines and obey traffic rules,
a pedestrian could move relatively freely on the open space.

Recently Zheng \emph{et.al. }\cite{hierarchical-trajecotry} proposed
to mediate such rigidity by dynamic re-evaluating goal along the way.
This method works with discrete grid-based scene structures with a
permanent set of goals such as basketball court; hence, it cannot
generalize to unseen scenes with arbitrary arrangements. Distinctively
from these approaches, we select destinations automatically from the
visual semantic features, and we dynamically learn from them using
the ranking and attention mechanism. By doing this, we enable our
model to be transferrable across different scenes.

\section{Method}

\begin{figure*}
\begin{centering}
\includegraphics[width=0.8\textwidth]{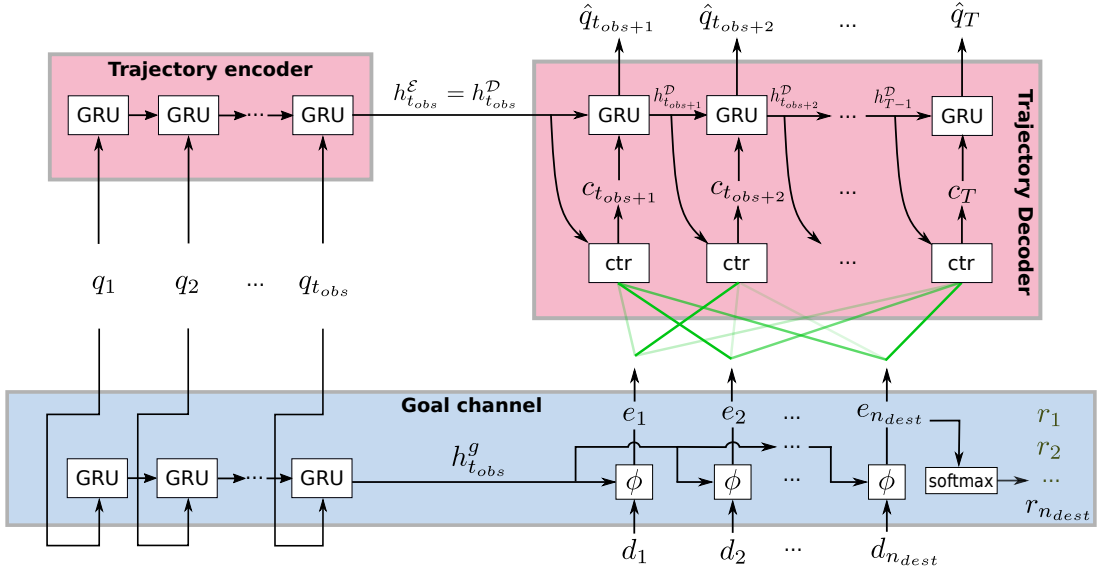} 
\par\end{centering}
\caption{\emph{Goal-driven trajectory prediction} architecture. The model contains
two neural channels: Goal channel (\textbf{\textcolor{blue}{blue}}
blocks) and Trajectory channel (\textbf{\textcolor{magenta}{pink}}
blocks). Goal channel matches the set of destinations $D=(d_{1},...,d_{n_{dest}})$
with the observed trajectory representation $h_{t_{obs}}^{g}$ and
modulates them into $E=(e_{1},...,e_{n_{dest}})$ using the supervision
of the goal ranks $R=(r_{1},...,r_{n_{dest}})$. In Trajectory channel,
the controlling signal $c_{t}$ is constructed at each time step by
attending to the modulated representation $E$ (\textbf{\textcolor{green}{green}}\textbf{
}lines). The weights of this attention mechanism are the compatibilities
between $E$ and the current hidden state $h_{t}^{\mathcal{D}}$ of
\emph{Trajectory decoder.} The signal $c_{t+1}$ then directs the
GRU units in \emph{Trajectory decoder} to predict the next position
$\hat{q}_{t+1}$(the additional feed-back input $\hat{q}_{t}$ of
this GRU is not drawn for clarity.)\label{fig:Model-architecture.} }
\end{figure*}

\subsection{Problem Definition}

We denote a person's trajectory from time $t=1$ to $t=T$ as $Q_{1:T}=(q_{1},q_{2},...,q_{T})$,
where $q_{t}=(x_{t},y_{t})$ is the 2D coordinate of the person at
time $t$. We also denote the observed trajectory of a person as $X=Q_{1:t_{obs}}$,
the future trajectory (ground truth) as $Y=Q_{t_{obs}+1:T}$ and the
predicted trajectory as $\hat{Y}=\hat{Q}_{t_{obs}+1:T}$.

We hypothesize that there are some underlying processes that govern
the trajectory $Q_{1:T}$; these processes reflect how a pedestrian
intends to reach a specific goal. In a 2D scene, the goal $g$ of
a pedestrian is defined to be the last visible point of the trajectory
that it usually lies at the border of the walkable area of the scene.
The set of all possible goals in a scene are gathered and clustered
into regions called destinations (green numbered regions in Fig. \ref{fig:intro-demonstration}),
where the number of destinations is $n_{dest}$. The continous goal
$g$ can be discretized to be one of these destinations: $g\in\left\{ 1,2,...,n_{dest}\right\} $.
In our work, the destinations are automatically detected; and each
of them is represented by an attribute vector $d_{i}\in\mathbb{R}^{6}$.
The detail of this representation is presented in Sec. \ref{par:Destination-representations-1}.

\subsection{Goal-driven Trajectory Prediction \label{subsec:Goal-conditioned-trajectory-pred}}

In this work, we want to study the relationship between the future
trajectory $Y$ and the goal $g$ of a pedestrian. These two terms
are strongly correlated but their behaviors are significantly different:
while people could quickly change their trajectories $Y$ adapting
to the surrounding, their goals $g$ usually remain stable throughout
the course of the movement. The joint distribution between them conditioned
on the observed part of the trajectory $X$ can be written as: 
\begin{equation}
P(Y,g|X)=P(Y|X,g)P(g|X).
\end{equation}
The future trajectory is obtained through marginalizing over all possible
goals:

\begin{equation}
P(Y|X)=\sum_{i}^{n_{dest}}P(Y|X,g=i)P(g=i|X).
\end{equation}
For this complex marginal distribution, we need to sample $Y$ for
every possible value of $g$, which could be computationally challenging.
To increase sampling efficiency, we use a mean-field style approximation:
\begin{align}
P(Y|X) & =\sum_{i}^{n_{dest}}P(Y|X,g)P(g=i|X)\approx P(Y|X,\bar{g}),\label{eq:approx_general}
\end{align}
where $\bar{g}$ is a continuous joint representation that is reflective
of the goal probability vector $P(g|X)$.

Building a meaningful $\bar{g}$ is crucial for the approximation
to work well; hence, this is at the center of our modeling. To this
end, we propose \emph{Goal-driven Trajectory Prediction} model (abbreviated
GTP) that explicitly characterizes the dependency of pedestrians'
future trajectory on their goal. The model consists of two channels,
each of which is a sub-network corresponding to one of the two hypothesized
sub-processes that control human trajectories. Among the two, Goal
channel matches the destinations by the observed trajectory and provides
the modulated destination representations to Trajectory channel. Then,
at each prediction time-step, Trajectory channel considers these representations
to calculate the adaptive goal vector $\bar{g}$ and use this vector
to forecast future movement. The overall architecture is demonstrated
in Figure \ref{fig:Model-architecture.}.

\paragraph{Goal channel}

The task of the Goal channel (blue block in Fig.\ref{fig:Model-architecture.})
is to observe the past trajectory $X=Q_{1:t_{obs}}$ and match it
with the destinations $D=(d_{1},...,d_{n_{dest}})$. It starts with
using a GRU unit to encode the observed signal:

\begin{equation}
h_{t}^{g}=\textrm{GRU(}h_{t-1}^{g},\gamma^{g}(q_{t})),
\end{equation}

where $h_{t}^{g}$ is the hidden state of GRU at time $t$, $\gamma^{g}$
is the function that embeds the position $q_{t}$ into a fixed length
vector. In this paper, we choose $\gamma^{g}$ to be a single layer
MLP.

After the observation period (at $t_{obs+1}$), the compatibility
between the hidden state $h_{t_{obs}}^{g}$ and each destination attribute
vector $d_{i}$ is measured through a joint representation:
\begin{align}
d'_{i} & =\text{MLP}(d_{i})\\
e_{i} & =\phi(\left[h_{t_{obs}}^{g},d'_{i}\right]),
\end{align}
where $d'_{i}$ is the embedded representation of destination $i$,
$\left[.,.\right]$ is concatenation operator and $\phi$ is the modulating
function chosen to be a single layer MLP with a tanh non-linearity
in our implementation.

The output $e_{i}$ is the\emph{ }representation\emph{ }of destination
$i$ modulated by the past trajectory. This representation captures
the pedestrian's perception of destination $i$ up to the point of
complete observation.

In order to make sure the network learns the compatibility, we force
$e_{i}$ to have the prediction power of the correct destination.
As the number of destinations varies across different scenes, we form
a ranking problem instead of standard classification alternatives:

\begin{equation}
r_{i}=\text{softmax}(\text{MLP}(e_{i})).\label{eq:ranking_score}
\end{equation}

The ranking is learned through the \emph{goal loss}: 
\begin{equation}
\mathcal{L}_{g}=-\log P(g=i^{*}|X)=-\log r_{i^{*}},\label{eq:goal_loss}
\end{equation}
where $g=i^{*}$ is the ground-truth goal.

After being ensured to contain goal-related information, the modulated
representations of destinations $E=(e_{1},...,e_{n_{dest}})$ are
provided to aid the Trajectory channel in predicting future trajectory
$Y$.

\paragraph{Trajectory channel. }

The trajectory channel (pink blocks in Fig.\ref{fig:Model-architecture.})
predicts future movements of the pedestrian by considering two factors:
the observed context $X=Q_{1:t_{obs}}$ and the modulated destination
representations $E=(e_{1},...,e_{n_{dest}})$. This channel is based
on a recurrent encoder-decoder network.

\emph{Trajectory encoder} is a GRU $\mathscr{\mathcal{E}}$ that takes
the observed trajectory $X=Q_{1:t_{obs}}$ and return the corresponding
hidden recurrent state, similarly as in the Goal channel:

\begin{equation}
\mathcal{E}:h_{t}^{\mathscr{\mathcal{E}}}=\text{GRU}(h_{t-1}^{\mathscr{\mathcal{E}}},\gamma^{\mathscr{\mathcal{E}}}(q_{t})),
\end{equation}

where $\gamma^{\mathscr{\mathcal{E}}}$ is an MLP with one layer.

\emph{Trajectory decoder }takes the role of generating future trajectories
$\hat{Y}$. It contains a GRU $\mathcal{D}$ that is innitialized
by the encoder's output $h_{t_{obs}}^{\mathcal{D}}=h_{t_{obs}}^{\mathscr{\mathcal{E}}}$and
recurrently rolls out future state $h_{t}^{\mathcal{D}},t=t_{obs+1,...,T}$.
This Trajectory Decoder stands out from traditional recurrent decoders
in that at each time step, it considers the modulated destinations
$E$ provided by the Goal channel as guidance in its prediction operations.

A straightforward solution to using $E=(e_{1},...,e_{n_{dest}})$
is by considering only $e_{i^{*}}$, which is the feature of the most
probable goal recognized by the Goal channel at the end of the observation
period. However, similar to the planning-based methods \cite{activity-forcasting},
this approach would be less flexible in the long term, as it does
not allow the choice of goals to be adjusted up to the situation.
Therefore, to maximize such adaptibility, we propose to dynamically
calculate a \emph{controlling signal} $c_{t}$ at each time step by
a specialized \emph{controller} sub-network abbreviated as \textbf{ctr:
\[
c_{t}=\boldsymbol{\textrm{ctr}}(E,h_{t-1}^{\mathcal{D}})
\]
}

Specifically, at time step $t>t_{obs}$, \textbf{ctr }takes the current
context $h_{t-1}^{\mathcal{D}}$ into account and reconsiders the
destinations attributes (modulated as $E=(e_{1},...,e_{n_{dest}})$)
through an attention mechanism. In detail, the attention weights $\alpha_{ti}$
on destination $i$ at time $t$ are calculated by matching $h_{t-1}^{\mathcal{D}}$
with destination modulated representation $e_{i}$:

\begin{equation}
\alpha_{ti}=\text{softmax}(\gamma^{a}(\left[e_{i},h_{t-1}^{\mathcal{D}}\right])),\label{eq:attention}
\end{equation}
where $\gamma^{a}$ is a single layer MLP with tanh activation.

Then, the controlling signal $c_{t}$ is computed softly from the
set of modulated representations:

\begin{equation}
c_{t}=\sum_{i=1}^{n_{dest}}\alpha_{ti}e_{i}.\label{eq:softgoal}
\end{equation}

Effectively, \textbf{ctr} builds the control signal $c_{t}$ by gathering
pieces of information from the options provided as destination modulated
representations. This process resembles an implementation of Random
Utility Theory \cite{random_utility_theory}, where the\textbf{ ctr}
block acts as a decision-maker that selects the \textquotedbl right\textquotedbl{}
choices from the set of alternatives, and attention weights $\alpha_{ti}$
plays as \emph{utility} factors. The control signal $c_{t+1}$ effectively
implements the goal vector $\bar{g}$ in Eq.\ref{eq:approx_general}.
It is combined with the previous prediction $\hat{q}_{t}$ to form
the input $s_{t+1}$ for the next step using another single layer
MLP embedding (not drawn in Fig.\ref{fig:Model-architecture.} for
clarity):
\[
s_{t}=\textrm{MLP}(\left[c_{t},\hat{q}_{t-1}\right]).
\]
The input $s_{t}$ is then fed into Trajectory decoder's GRU $\mathcal{D}$
to roll toward the future: 

\[
\mathcal{D}:h_{t}^{\mathcal{D}}=\text{GRU}(h_{t-1}^{\mathcal{D}},s_{t}).
\]
After each rolling, the predicted output $\hat{q_{t}}$ is generated
by the output MLP $\gamma^{o}$:

\begin{equation}
\hat{q}_{t}=\text{\ensuremath{\gamma^{o}}}(h_{t}^{\mathcal{D}}).
\end{equation}

The loss function in Trajectory channel is simply the distance between
the predicted trajectory $\hat{Y}$ and the ground truth $Y$: 
\begin{equation}
\mathcal{L}_{tr}=\frac{1}{T}||Y-\hat{Y}||_{2},\label{eq:traj_loss}
\end{equation}

where $T$ is the prediction length and $\mathcal{L}_{tr}$ is the
Trajectory loss.

\subsection{Destination Selection}

\begin{figure}
\begin{centering}
\includegraphics[width=0.49\textwidth,height=5cm,keepaspectratio]{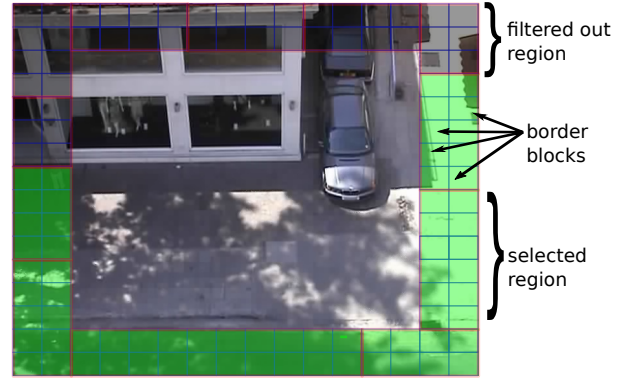}
\par\end{centering}
\caption{Destination selection. We first divide the scene into multiple blocks
(blue grid). Then, we group border blocks into regions using semantic
feature similarities (red rectangles). The regions with high walkable
scores are then selected as destinations for GTP (green filled regions).
\label{fig:dest}}
\end{figure}

An important preprocessing task supporting GTP is automatically constructing
a good set of destinations and extract their meaningful features $D=(d_{1},...,d_{n_{dest}})$
from the scene. These destinations must cover most possible final
points of the trajectories, and in the meantime reflect accurately
options for pedestrians' chosen goals. 

Our destination selection process includes five steps, and it begins
by extracting the static background scene $B$ from the video frames.
An example of $B$ is shown as background of Fig. \ref{fig:dest}.
Background $B$ is then segmented into semantic areas with the Cascade
Segmentation Module \cite{pretrained_seg1} trained on ADE20K dataset
\cite{pretrained_seg2}:
\[
S,F=\textrm{semantic\_parse}(B)
\]
where $S$ contains the scores of segmentation and $F$ is the feature
map extracted from the penultimate layer of the model. Both tensors
have the same spatial size as the image; $S$ has the depth $150$
corresponding to categories and features in $F$ has the length of
$512$.

Then, as GTP works on a small discrete set of destinations, we divide
a scene into $N\times N$  blocks. Among these blocks, we only consider
those at the boundary and the ones next to them. This set of blocks
is called ``border blocks'', and it is drawn in Fig. \ref{fig:dest}. 

In the third and fourth steps, we compute the semantic feature of
each border block by average pooling features $F$ from its pixels.
We then cluster nearby border blocks into regions based on the similarity
between their features. At the end of these steps, we have a set of
connected regions potentially be destinations for GTP.

Finally, to exclude regions that cannot be realistic destinations,
we further filter out the ones of ``non-walkable'' categories by
using score tensor $S$. For each region, the maximum scores of the
walkable categories (selected from scene labels) is compared against
a threshold to select the final destination regions $D$. For each
of these regions, its feature $d_{i}$ is calculated by agent-centric
geometrical measure detailed in the next section.

\subsection{Agent-centric Representation\label{par:Destination-representations-1}}

\begin{figure}
\begin{centering}
\includegraphics[width=0.49\textwidth,height=5cm,keepaspectratio]{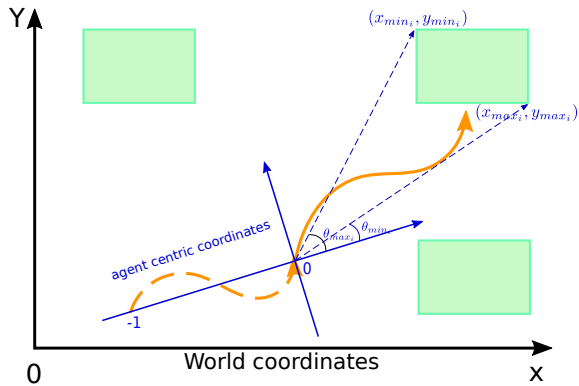}
\par\end{centering}
\caption{The agent-centric coordinates is defined to have the root at the agent's
location at the prediction time and directed toward the overall past
direction. Destination features are represented as the combination
of the relative distances and angles. \label{fig:Agent-centric-coordinates:-(left}}
\end{figure}

\begin{figure*}
\begin{minipage}[t]{0.49\textwidth}%
\begin{center}
\includegraphics[width=0.89\textwidth]{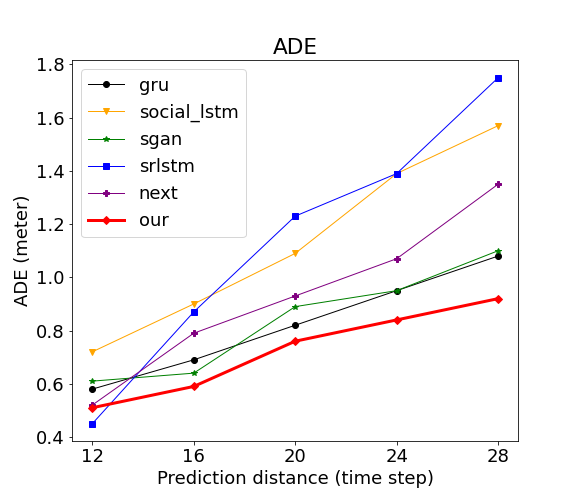}
\par\end{center}%
\end{minipage}\hfill{}%
\begin{minipage}[t]{0.49\textwidth}%
\begin{center}
\includegraphics[width=0.89\textwidth]{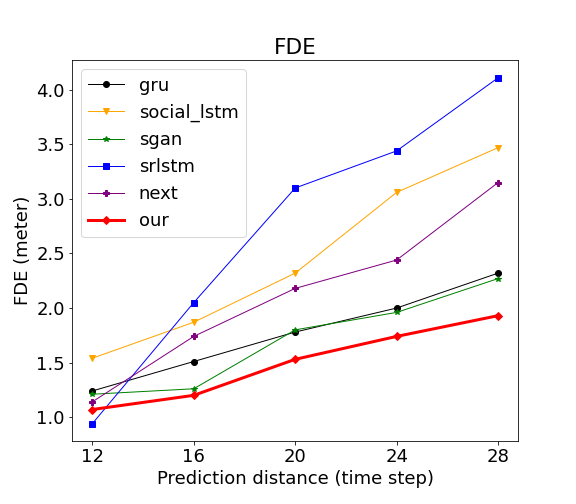}
\par\end{center}%
\end{minipage}

\caption{Performance of compared models in ADE - left and FDE - right (the
lower the better) on ranging prediction length. $\protect\Model$
(\textbf{\textcolor{red}{red}} diamond) shows increasingly favorable
performance over baselines while other state-of-the-art only have
the advantage in short-range prediction. Detailed numeric results
are included in the supplementary materials. \label{fig:adefde}}
\end{figure*}

Different from other RNN based future predictors \cite{sophie,sr-lstm},
both of GTP's channels rely heavily on the personalized perception
of each pedestrian about the destinations. Although the network could
learn to adapt the geometrical relationship in any reference frames,
we want it to concentrate on modeling the relative perception of goals
and trajectory rather than the absolute scale of the scenes. For that
purpose, we represent both the trajectory and the destinations' geometrical
features in the personalized coordinate system with respect to each
pedestrian called \emph{the agent-centric coordinate} (see Fig. \ref{fig:Agent-centric-coordinates:-(left}). 

These coordinates is defined to be rooted at $q_{t_{obs}}$ and has
the unit vector $\overrightarrow{u}=(-1,0)$ transformed from $\overrightarrow{q_{t_{obs}}q_{1}}$.
All parts of the trajectory including observed $X$ and predicted
$Y$ are transformed to this system before being used in GTP.

Under the agent-centric coordinate, the destination features $d_{i}$
includes the relative distance and direction of the destination in
the perspective of the pedestrian:
\[
d_{i}=(x_{min_{i}},y_{min_{i}},x_{max_{i}},y_{max_{i}},\theta_{min_{i}},\theta_{max_{i}}),
\]
where the first four elements are the coordinate of the destination
regions and the last two represent angle ranges from pedestrian's
point of view.

Experimental results showing the effective of this representation
is detailed in the ablation study (Sec. \ref{subsec:Ablation})

\subsection{Training GTP}

When training the model, we need to attain the balance and collaboration
between optimizing the Goal channel loss $\mathcal{L}_{g}$ and the
Trajectory channel loss $\mathcal{L}_{tr}$. To this end, we use a
three-stage training process. At the first stage, we fix the Trajectory
channel and only train Goal channel with $\mathcal{L}_{g}$ to ensure
that the modulated destinations $E$ capture goal-related information.
Then, in the second stage, we freeze the Goal channel, keeping the
modulated representations $E$ unchanged, and train Trajectory channel
with $\mathcal{L}_{tr}$. Finally, in the last stage, we refine the
whole model using $\mathcal{L}_{tr}$.

\section{Experiments}

\begin{figure*}[!t]
\noindent %
\noindent\begin{minipage}[t]{1\textwidth}%
\includegraphics[width=0.6\textwidth]{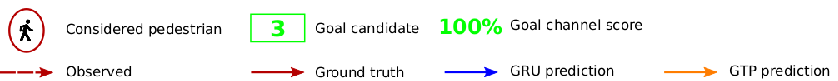}%
\end{minipage}

\vfill{}

\subfloat[\label{fig:sucessa}]{\includegraphics[width=0.45\textwidth]{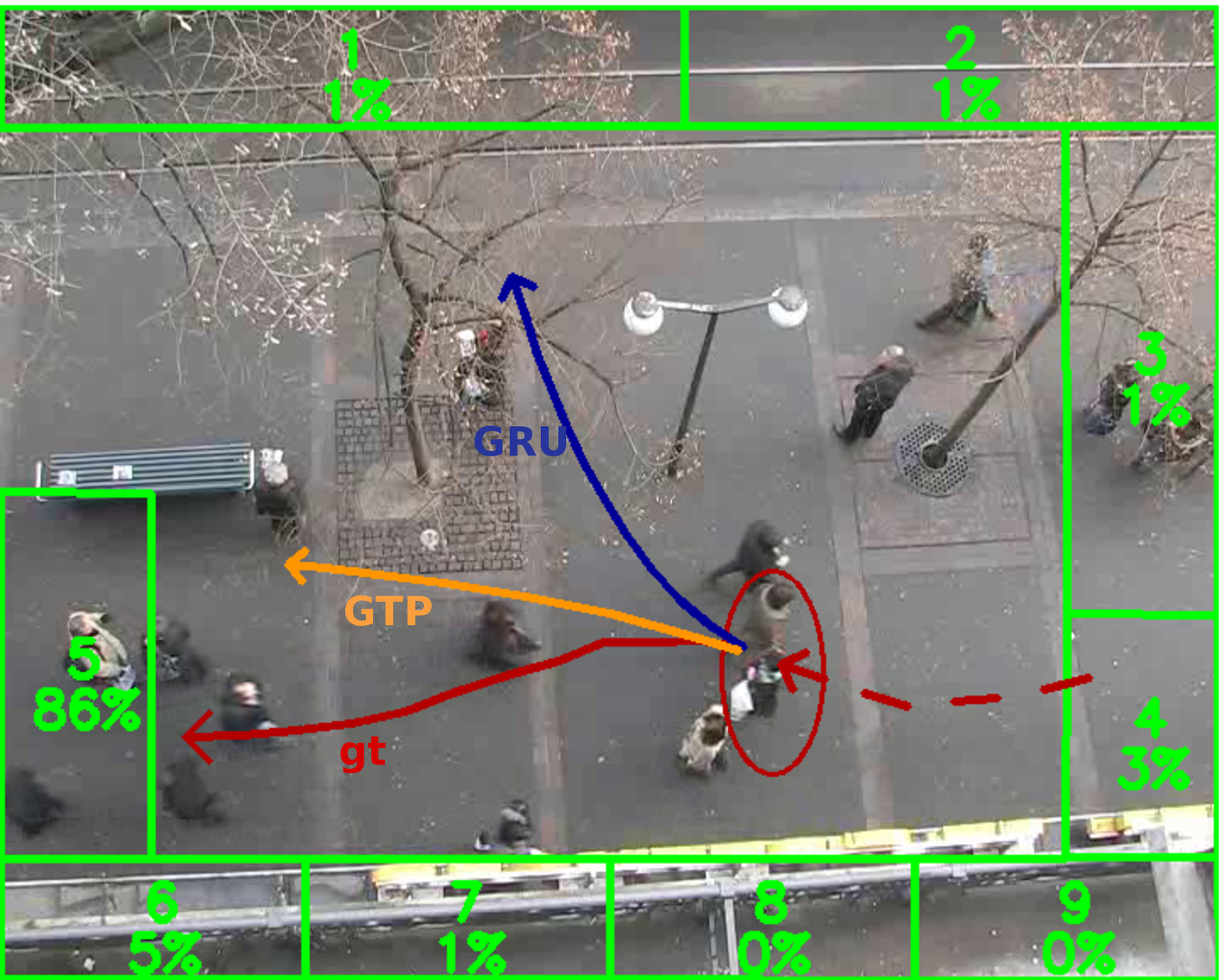}

}\hfill{}\subfloat[\label{fig:fig:sucessb}]{\includegraphics[width=0.45\textwidth]{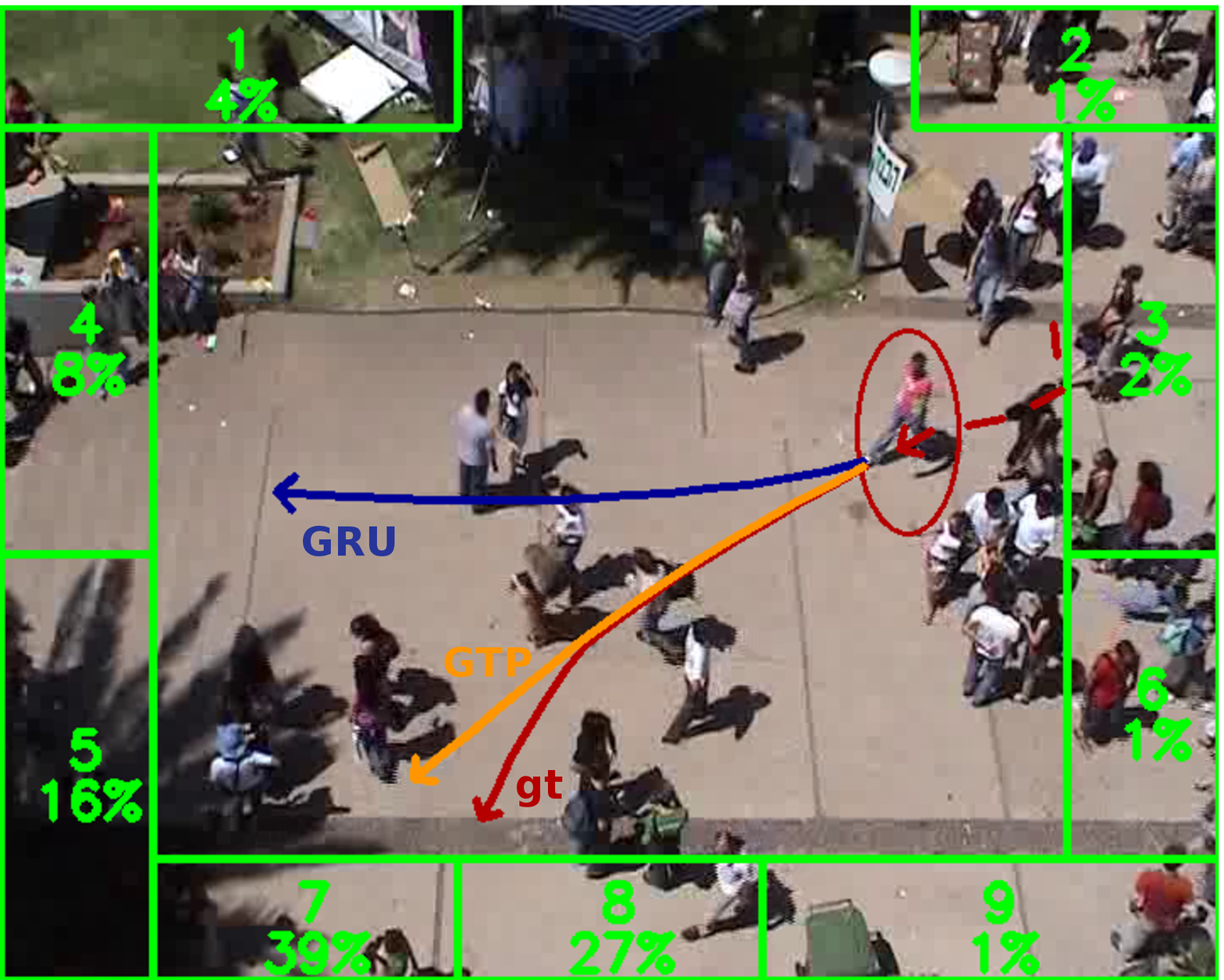}

}

\caption{Qualitative analysis of $\protect\Model$. \textbf{(a) }GRU predicts
the pedestrian to follow the upward trend in the observed trajectory,
while\textbf{ }$\protect\Model$ estimates the destination to be Goal
5 and generates trajectory toward that destination. \textbf{(b) }Similarly,
the correct estimated goal (Goal 7) enable $\protect\Model$ to perform
better than simple GRU. We provide more cases in supplementary materials.
\label{fig:success}}
\end{figure*}

\subsection{Experiment Settings}

\paragraph{Dataset}

We use the two most prominent benchmark datasets for trajectory predictions
which are ETH \cite{eth} and UCY \cite{ucy}. These datasets contain
real-world annotated trajectories in five different scenes: ETH, HOTEL
(from ETH dataset), ZARA1, ZARA2, UNIV (from UCY dataset). Similar
to previous works \cite{sociallstm,socialgan,peekingintofuture,sr-lstm},
we preprocess the trajectories to convert from camera coordinates
to world coordinates using provided homography parameters.

\paragraph{Baselines}

We compare our method against four state-of-the-art methods Social
LSTM \cite{sociallstm}, Social GAN \cite{socialgan}, SR-LSTM \cite{sr-lstm}
and Next \cite{peekingintofuture}. We also use a baseline GRU Encoder-Decoder,
which is the base architecture that all of these works developed on
top of.\footnote{Several methods use LSTM units instead of GRU in the encoder-decoder
architecture. The two units are fundamentally similar and our experiments
showed that they achieve identical results, therefore we only use
GRU units in the baseline for the sake of concision.}

\paragraph{Evaluation protocols and metrics}

The common setting used in the state-of-the-arts is observing 8 time
steps (3.2 seconds) and predict the near-term future trajectories
in the next 12 time steps (4.8 seconds). With the objective of long-term
prediction, we extend this setting into maximal future distance possible
supported by the dataset. This results in the settings of observing
8 and predicting 12, 16, 20, 24, 28 time steps. For all of the methods,
a model for each of these settings is trained separately.

As with other methods, we use the leave-one-out cross-validation approach,
training on 4 scenes, and testing on the remaining unseen scene. The
performance is measured using two error metrics: Average displacement
error (ADE) and Final displacement error (FDE).

In reporting results, we cite the results reported in the baseline
papers when possible. In the new settings of far-term prediction,
we retrain these models using the standard published implementation.

\subsection{Quantitative Performances \label{subsec:Quantitative-Evaluation}}

The performance of GTP and other methods are graphed in Figure \ref{fig:adefde}.
It showed that some of the state-of-the-art methods, namely SR-LSTM
\cite{sr-lstm} and Next \cite{peekingintofuture} have advantages
over baseline GRU only in the short-term 12 time-step prediction.
However, all of these methods have fast deteriorating performances
along with prediction distances.

By contrast, $\Model$'s performances are much more stable in the
far-term prediction. Although the model is slightly worse than SR-LSTM
in 12 time steps prediction, it outperforms all of the baselines in
other prediction settings. Also, the farther the prediction is, the
gap between GTP and other methods getting more and more widened.

The strong performance of $\Model$ in far-term prediction could be
attributed to the collaborative interaction between the two channels.
By considering the modulated representations $E=(e_{1},...,e_{n_{dest}})$
provided by Goal Channel, Trajectory Channel has access to the possible
goals of the pedestrian; therefore, the model could generate consistent
future locations toward the correct destination. By contrast, baseline
models are unaware of the goal concept and only use the immediate
continuity of the trajectory. Without the long-term guidance, the
recurrent predictors accumulate the errors in each step, leading to
plummeted performance.

\subsection{Qualitative Analysis}

We further investigate the dependence of future trajectories on goals
by visualizing the results generated from $\Model$ and GRU baseline.
As shown in Figure \ref{fig:success}, GRU could only predict future
trajectories based on the dynamic of the observed trajectory, and
hence it will fail when the moving pattern is complex. Certainly,
in Figure \ref{fig:sucessa}, GRU predicts the pedestrian to follow
his upward trend, while the true trajectory is to move forward. By
contrast, $\Model$ predicts that the pedestrian will reach Goal 5
and generate future positions toward that destination. Similarly,
in Figure \ref{fig:fig:sucessb}, the estimated goal enable $\Model$
to predict accurate trajectory, which could not be achieved in simple
GRU.

\begin{table*}
\begin{centering}
\begin{tabular}{|l|l|c|c|c|c|c|}
\hline 
\# &  & obs8 - pred12 & obs8 - pred16 & obs8 - pred20 & obs8 - pred24 & obs8 - pred28\tabularnewline
\hline 
\hline 
0 & Full GTP model & 0.51 / 1.07 & \textbf{0.59 / 1.20} & \textbf{0.76 / 1.53} & \textbf{0.84 / 1.74} & \textbf{0.92 / 1.93}\tabularnewline
\hline 
1 & W/o destination modulation & 0.52 / 1.09 & 0.65 / 1.37 & 0.86 / 1.82 & 1.23 / 2.67 & 1.24 / 2.77\tabularnewline
\hline 
2 & W/o goal loss & \textbf{0.50 / 1.06} & 0.60 / 1.24 & 0.78 / 1.61 & 0.92 / 1.94 & 1.07 / 2.30\tabularnewline
\hline 
3 & W/o flexible attention weights & 0.51 / 1.07 & 0.62 / 1.3 & 0.76 / 1.57 & 0.88 / 1.84 & 0.99 / 2.09\tabularnewline
\hline 
4 & W/o agent-centric coordinates & 0.6 / 1.28 & 0.68 / 1.46 & 0.81 / 1.76 & 0.96 / 2.05 & 1.01 / 2.12\tabularnewline
\hline 
5 & W/o any goal-driven features & 0.58 / 1.24 & 0.69 / 1.51 & 0.82 / 1.78 & 0.95 / 2.0 & 1.08 / 2.32\tabularnewline
\hline 
\end{tabular}
\par\end{centering}
\caption{Ablation experiments for trajectory prediction in ADE/FDE. The lower
the numbers, the better the model. \label{tab:tab3}}
\end{table*}

\subsection{Model Behavior Analysis}

\begin{figure}
\subfloat[\label{fig:attention1}]{\begin{centering}
\begin{minipage}[t]{0.22\textwidth}%
\includegraphics[width=1\textwidth,height=4cm,keepaspectratio]{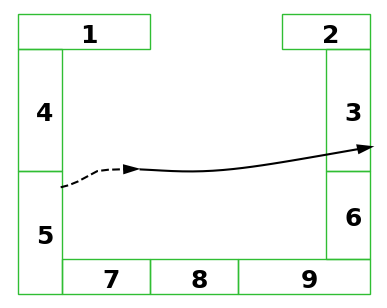}%
\end{minipage}\hfill{}%
\begin{minipage}[t]{0.22\textwidth}%
\includegraphics[width=1\textwidth,height=3cm,keepaspectratio]{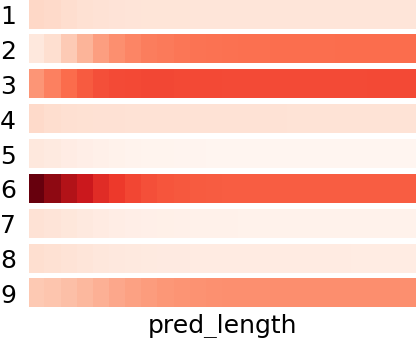}%
\end{minipage}
\par\end{centering}
}\vfill{}
 \subfloat[\label{fig:attention2}]{%
\begin{minipage}[t]{0.22\textwidth}%
\includegraphics[width=1\textwidth,height=4cm,keepaspectratio]{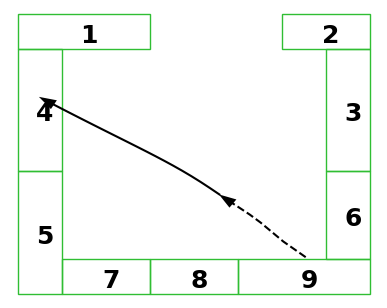}%
\end{minipage}\hfill{}%
\begin{minipage}[t]{0.22\textwidth}%
\includegraphics[width=1\textwidth,height=3cm,keepaspectratio]{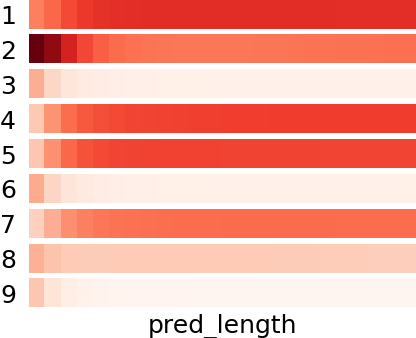}%
\end{minipage}\centering{}

}

\caption{Patterns of attention weights used in \textbf{ctr} block as trajectory
prediction progresses. The farther toward the end, the sharper attention
was put on narrowed down choices among the destinations. \label{fig:attention}}
\end{figure}

To have a better view of how the Trajectory channel uses the modulated
destinations $E$, we visualize the attention weights $\alpha_{ti}$
in \textbf{ctr }block in several cases (see Figure \ref{fig:attention}).

In the first example (Fig. \ref{fig:attention1}), the \textbf{ctr
}block initially paid the most attention to destination 3 and 6 and
gave uniform weights to the rest. As the prediction progressed, it
gradually concentrated more on destination 2 and 9 as they became
potential goal choices.

In the second example (Fig. \ref{fig:attention2}), at the early stage,
\textbf{ctr} considered destination 2 the most. However, at the later
stage, as the trajectory progressed from right to left, destination
2 is no longer the most potential candidate, it switched to considering
the information contained in the potential goals in destinations 1,
4, 5, and 7. The attention weights of these short-listed destinations
got sharper and sharper attention as the probable goal choices narrowed. 

\subsection{Ablation Study \label{subsec:Ablation}}

We provide more insight about the roles of $\Model$ components and
choices by turning off each of the key proposed components. The results
of the study are reported in Table \ref{tab:tab3}.
\begin{enumerate}
\item \textbf{Without destination modulation}. We investigate the contribution
of modulated representations by ablating them from the model. Specifically,
in the third row of Table \ref{tab:tab3}, we use the raw destination
representations $D$, instead of the modulated representations $E$,
to compute the controlling signal in Eq. \ref{eq:attention}, \ref{eq:softgoal}.
The results show that using raw destination features is better than
not using them at all (row 1 vs row 5). However, modulating those
features by the Goal channel helps GTP reaches its maximum advantages
(row 0).
\item \textbf{Without goal loss. }Without using the goal loss, there is
no guarantee that the modulated representations contain goal-related
information. Consequently, the performances of GTP in long-term prediction
(20 to 28 time-steps) decrease significantly.
\item \textbf{Without flexible attention weights.} In this experiment, we
test the straightforward alternative of directly using the ranking
scores $r_{i}$ of the Goal channel (Eq. \ref{eq:ranking_score})
in place of $\alpha_{ti}$ to compute the controlling signal $c_{t}$
in Eq. \ref{eq:softgoal}; by doing this, we bypass the computation
of attention weights $\alpha_{ti}$ at every time step. Although saving
some computation, this rigid option does not allow attention to adapt
to the evolving situation; hence, it leads to a reduction in performance,
especially in far-terms. 
\item \textbf{Without agent centric coordinates. }Using the raw world coordinates
hurts the performance of $\Model$ significantly. This indicates that
agent-centric coordinate plays an important role in estimating and
utilizing agent's goal information.
\item \textbf{Without any goal-driven features. }This model is equivalent
to a vanilla GRU encoder-decoder baseline. It is clear that most of
the proposed features make improvements over this baseline; and combination
of these features in the full GTP model leads to the best result.
\end{enumerate}

\section{Conclusion}

In this work, we study the problem of forecasting human trajectory
into the far future. We propose GTP: a dual-channel network for jointly
modeling the goals and trajectory of pedestrians. Through experiments,
we have verified that GTP effectively takes advantage of the imposed
goal structures and provides strong consistency in long-term prediction,
and reduces the progressive accumulation of error. 

This work opens room for future investigations on considering long-term
goals together with the short-term social and environmental context.
Static environmental factors can be incorporated easily into the input
at each step. Social interaction can be introduced through message
passing between trajectory channels and shared goal modeling.

{\small{}\bibliographystyle{ieee_fullname}
\bibliography{egbib}
}{\small\par}
\end{document}